\def\year{2021}\relax
\documentclass[letterpaper]{article} 
\usepackage{aaai21}  
\usepackage{times}  
\usepackage{helvet} 
\usepackage{courier}  
\usepackage[hyphens]{url}  
\usepackage{graphicx} 
\usepackage{natbib}  
\usepackage{caption} 
\usepackage{booktabs}
\usepackage{amsmath}
\usepackage{amssymb}
\usepackage{algorithm}
\usepackage{algorithmic}
\usepackage{array}
\usepackage{subfig}

\title{Spatial-Temporal Dynamic Graph Attention Networks for Ride-hailing Demand Prediction}
\author{Weiguo Pian\textsuperscript{\rm 1}, Yingbo Wu\thanks{Corresponding author}\textsuperscript{\rm 2}, Xiangmou Qu\textsuperscript{\rm 2}, Junpeng Cai\textsuperscript{\rm 2}, Ziyi Kou\textsuperscript{\rm 3}\\ 
\textsuperscript{\rm 1}Automotive Collaborative Innovation Center, Chongqing University\\ 
\textsuperscript{\rm 2}School of Big Data \& Software Engineering, Chongqing University\\
\textsuperscript{\rm 3}Department of Computer Science and Engineering, University of Notre Dame\\
\{pwg,wyb,qxm,cjp\}@cqu.edu.cn, zkou@nd.edu
}
\begin{document}

\maketitle

\thispagestyle{empty}

\begin{abstract}
Ride-hailing demand prediction is an essential task in spatial-temporal data mining. Accurate Ride-hailing demand prediction can help to pre-allocate resources, improve vehicle utilization and user experiences. Graph Convolutional Networks (GCN) is commonly used to model the complicated irregular non-Euclidean spatial correlations. However, existing GCN-based ride-hailing demand prediction methods only assign the same importance to different neighbor regions, and maintain a fixed graph structure with static spatial relationships throughout the timeline when extracting the irregular non-Euclidean spatial correlations. In this paper, we propose the Spatial-Temporal Dynamic Graph Attention Network (STDGAT), a novel ride-hailing demand prediction method. Based on the attention mechanism of GAT, STDGAT extracts different pair-wise correlations to achieve the adaptive importance allocation for different neighbor regions. Moreover, in STDGAT, we design a novel time-specific commuting-based graph attention mode to construct a dynamic graph structure for capturing the dynamic time-specific spatial relationships throughout the timeline. Extensive experiments are conducted on a real-world ride-hailing demand dataset, and the experimental results demonstrate the significant improvement of our method on three evaluation metrics \textit{RMSE}, \textit{MAPE} and \textit{MAE} over state-of-the-art baselines.
\end{abstract}

\section{Introduction}\label{sec1}
With the rapid development of mobile internet and sharing economy around the world, ride-hailing has become more and more popular in recent years. Online Ride-hailing platforms, such as Didi Chuxing, Uber, and Grab, provide a convenient way of traveling for people.


However, due to the scheduling and allocation of cars vary a lot depending on specific user requirements, these platforms still suffer from some inefficient operations~\cite{taxi1}, such as long passenger waiting time~\cite{taxi2} and excessive vacant trips~\cite{taxi3}. Therefore, an accurate ride-hailing demand prediction method can help organize vehicle fleet, improve vehicle utilization, reduce the wait-time, and mitigate traffic congestion~\cite{AAAI2019}.


To tackle this problem, there have been several methods proposed in recent years focusing on similar prediction tasks, including traffic volume, taxi pick-ups, and traffic in/out flow volume prediction. Traditional methods applied statistical models to solve these problems~\cite{GaussianMixtureModel,DemandStreamingData,AdaptiveSeasonal}. Recently, deep learning methods have been widely used in these tasks. One of the first deep learning methods in these spatial-temporal prediction problems was introduced by Zhang et al.~\cite{DeepST} who applied deep convolutional layers for prediction. After that, Wang et al. concatenated several related factors as inputs to predict the gap between taxi supply and demand via a non-linear MLP network~\cite{DeepSD}. Based on~\cite{DeepST}, Zhang et al.~\cite{AAAI2017} further proposed a deep convolutional network named ST-ResNet to predict in-out traffic flow among different areas. However, both of them did not consider the temporal information hidden in the sequential data which is an important factor in transportation issues. Based on that, Yao et al.~\cite{AAAI2018} constructed a spatial-temporal model to predict various taxi demands, and they further created a graph embedding module to pass information among different regions. Though they capture the topological information, it is hard to aggregate demand values from related regions through the region-wise relationship. Besides, applying convolutional neural networks can only model regular Euclidean correlations among different regions. To capture the irregular non-Euclidean pair-wise correlations, Yu et al.~\cite{STGCN} applied GCN to model the spatial relationships for traffic prediction. After that, Geng et al.~\cite{AAAI2019} proposed a muti-GCN-based model to forecast the ride-hailing demand. However, when extracting spatial features among nodes (regions), traditional GCN can only assign the same importance to their neighbor nodes (regions)~\cite{GAT}, which resulted in a serious lack of adaptive weights allocation for different neighbors to model the adaptive pair-wise correlations.

\begin{figure}[t]
    \centering
    \includegraphics[width=.5\textwidth]{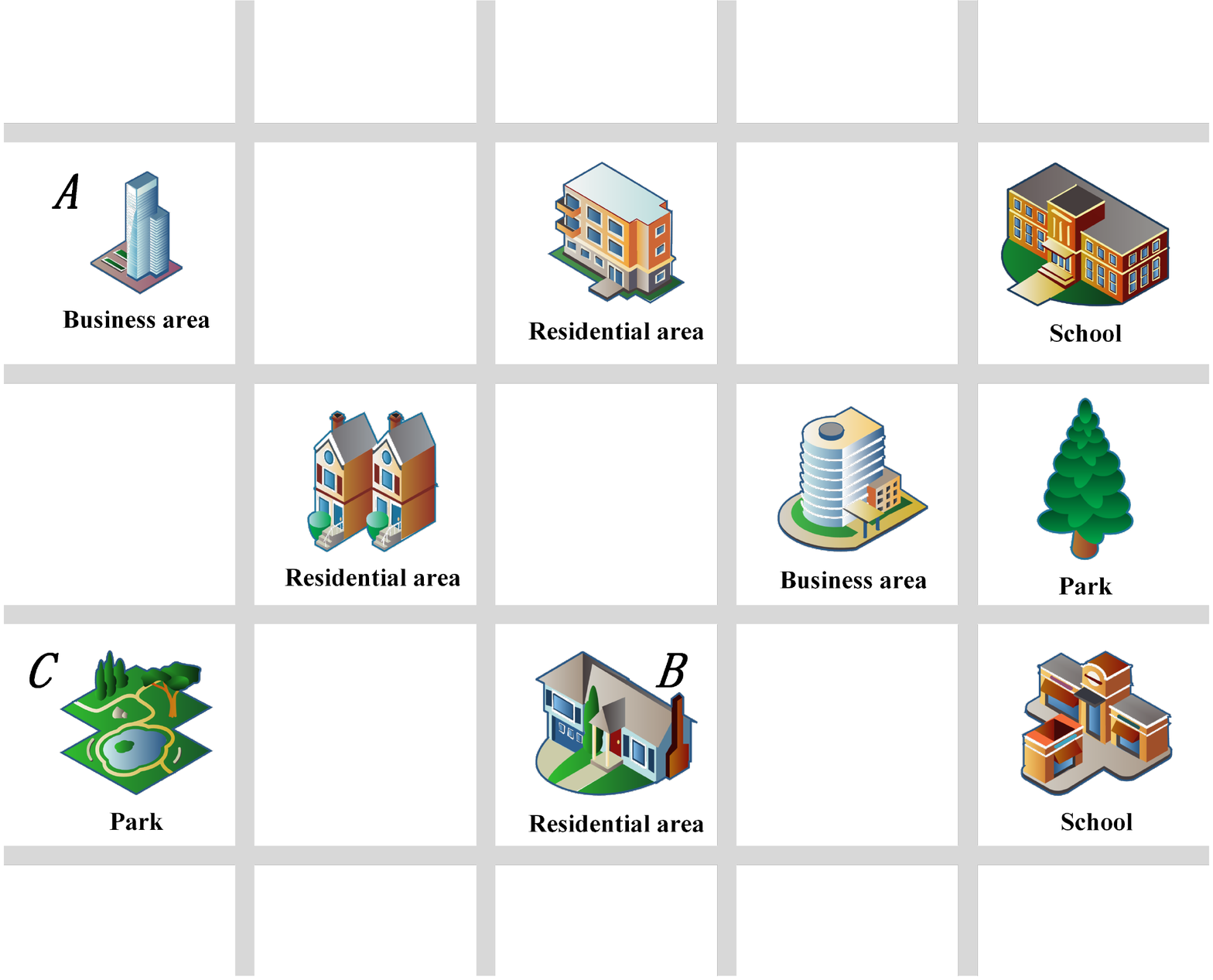}
    \caption{An example of different regions. Region $A$, $B$ and $C$ are business area, residential area and park respectively.}
    \label{fig:map}
\end{figure}

In this paper we propose a novel deep learning method, called Spatial-Temporal Dynamic Graph Attention Networks (STDGAT). In the spatial module of STDGAT, we utilize Graph Attention Network (GAT)~\cite{GAT} to extract non-Euclidean spatial correlations among different regions. The GAT is an advanced graph neural network, in which the nodes (regions) can allocate different importance for their neighbors adaptively based on attention mechanisms~\cite{Attention}. In this way, the advantage of GAT can be fully utilized to model the different pair-wise correlations for the different neighbor regions.

In addition, the connection of nodes in the graph should not be based on real geographical regions connection which is fixed throughout the timeline. For example, as figure~\ref{fig:map} shows, region $A$ and $B$ are business areas and residential areas respectively. Region $A$ may be affected by $B$ in the daytime while there may be not many correlations between region $A$ and $B$ in the evening. 

To address this issue, STDGAT adopts the time-specific commuting-based graph attention mode. For each time interval, the edges between nodes (regions) pairs are based on their actual commuting relationships instead of geographical connections. For example, in figure~\ref{fig:map}, if there do not have any commuting from region $B$ to $C$ but have some commuting from region $C$ to $B$ at time interval $t$, the edge from node $B$ to $C$ is not existing while that from node $C$ to $B$ is existing at time interval $t$. In this way, each node can perform aggregation operations for the nodes with actual commuting relationships in time interval $t$, which achieves the dynamic neighbor nodes selection.



Long Short-Term Memory (LSTM) has been proven to have excellent performances in temporal dependency modeling\cite{AAAI2018,Origin-Destination-Demand}. Therefore, in our temporal module of STDGAT, we apply the LSTM~\cite{LSTM} to model the temporal dependencies among the sequential output from the spatial module. In this way, STDGAT can extract the the temporal dependency features among sequences for prediction.

We conduct extensive experiments to validate our STDGAT method on a large-scale real-world ride-hailing dataset from Didi Chuxing, one of the largest ride-hailing platforms in China. The dataset contains ride-hailing orders of Didi Chuxing in Haikou, China, from May 1st to October 31st with the total number of 12,185,427 orders, which can be downloaded from the website of Didi Chuxing GAIA Initiative\footnote{https://gaia.didichuxing.com}. The results of the experiments demonstrate the superiority of our STDGAT over state-of-the-art methods.

Our contributions are highlighted as following:
\begin{itemize}
    \item We propose a novel ride-hailing demand prediction method STDGAT which is constructed by a spatial module and a temporal module. In the spatial module of STDGAT, based on the attention mechanism of GAT, we extract different pair-wise correlations by adaptive importance allocation to model the non-Euclidean spatial correlations for different neighbor regions. In the temporal module of STDGAT, we use LSTM to model the temporal dependencies among the sequential outputs from the spatial module.
    \item By adopting the time-specific commuting-based graph attention mode to construct  dynamic graph structures, STDGAT can capture the dynamic time-speciﬁc spatial relationships at different time intervals throughout the timeline, so as to achieve the dynamic neighbor nodes selection for each node (region).
    \item We conducted extensive experiments on a large-scale real-world ride-hailing dataset from Didi Chuxing. The experimental results show the superiorities of our method on three evaluation metrics \textit{RMSE}, \textit{MAPE} and \textit{MAE}, compared with state-of-the-art baselines.
\end{itemize}

The remainder of this paper is organized as follows. Section II introduces some related works by describing some recent papers. Section III fixes some notations and formulates our demand prediction problem mathematically. Section IV describes the details of our proposed STDGAT. Section V compares STDGAT with state-of-the-art methods and conducts some ablation studies on the real-world ride-hailing dataset from Didi Chuxing. Finally, Section VI concludes this paper and discusses the future work.

\section{Related Work}\label{sec2}
\subsection{Spatial-Temporal prediction in urban computing}
Spatial-temporal prediction is one of the main problems in urban computing, which includes the tasks of traffic flow prediction, destination prediction, demand prediction (our task), etc. These tasks are kind of similar. Essentially, they use the historical sequence data (time series data) to predict the future data~\cite{DeepSD,AAAI2018,AAAI2017,DeepST}. Some traditional methods only rely on time series information to regress the prediction results. One of the most representative methods is Autoregressive Integrated Moving Average (ARIMA) that has been widely used in time series prediction~\cite{DemandStreamingData, AdaptiveSeasonal}. To further improve the model's performance, some other data, such as weather conditions and event information, has also been considered as external context information~\cite{realworld, UbiquitousUrbanData}. Recently, with the significant progress in deep learning, some researchers began to focus on deep learning to solve such problems. One of the first DNN-based crowd flow prediction methods was proposed by Zhang et al.~\cite{DeepST}. After that, inspired by ResNet~\cite{ResidualNet}, they further applied residual connection to their model and proposed the ST-ResNet~\cite{AAAI2017}. Wang et al.~\cite{DeepSD} used many context data as inputs of their model to predict the region-level supply-demand gap of taxi. Moreover, some RNN-based methods were proposed to utilize the temporal dependencies hidden in the data~\cite{ExtremeConditionTrafficForecasting,Destination-sub-trajectory}. To extract the joint spatial-temporal feature of the data, Yao et al.~\cite{AAAI2018} combined CNN and LSTM into a single model, and used a fixed graph embedding to extract the constant feature among regions. Qiu et al.~\cite{Origin-Destination-Demand} proposed a contextualized spatial-temporal network for taxi origin-destination demand prediction. Though they achieved great success in some spatial-temporal prediction fields, they neglect the modeling of non-Euclidean spatial correlations among regions. To overcome this problem, Li et al.~\cite{DCRNN} proposed a graph convolution-based recurrent neural network for traffic forecasting. Besides, Yu et al.~\cite{STGCN} proposed ST-GCN, which used graph convolutional networks to extract the non-Euclidean spatial information and 1-D convolution for temporal information modeling, for traffic prediction. After that, Geng et al.~\cite{AAAI2019} constructed muti-graphs, i.e., neighborhood, functional similarity, and transportation connectivity, for ride-hailing demand prediction, and they applied GCN and RNN to capture the spatial and temporal features respectively.

However, all the previous methods we mentioned above lack of modeling different spatial correlations for different neighbor regions, and they ignored the importance of constructing region graphs dynamically throughout the timeline.

\subsection{Deep Learning}
Deep learning has been successfully used in a large number of fields, especially Convolutional Neural Networks (CNN) in the field of computer vision~\cite{ImageNet, AAAI2017}. The proposed of Residual Network enables the neural networks to reach an unprecedented depth~\cite{ResidualNet}.  
In recent years, Recurrent Neural Networks (RNN) have achieved good results in sequence learning tasks~\cite{seq2seq}. The incorporation of Long Short-Term Memory (LSTM) overcome the shortage of traditional RNN that the difficulty of learning long-term dependency~\cite{GradientFlowinRecurrentNets,AAAI2017}. However, these networks can only model for spatial or temporal features. Xingjian et al.~\cite{ConvLstm} propose a model named convolutional LSTM network which combines the characteristics of CNN and LSTM to model for spatial-temporal features. However, this model is difficult to model long-range temporal dependencies and will become more difficult to train as the network deepens~\cite{AAAI2017}.

Recently, Graph Neural Networks (GNN) have become an important subfield of deep learning, which has been widely used in many fields, such as social networks~\cite{GCN_Niepert}, computer vision~\cite{RGCN_Action_Forecasting,STGCN_Action_Recognition}, and traffic prediction~\cite{STGCN, AAAI2019}. Traditional Convolutional Neural Networks (CNN) can only extract features from standard grid data. However, many kinds of data can not be represented in a grid-like structure~\cite{GAT}. Therefore, GNNs were proposed to process such kind of data. The earliest GNNs were introduced by Gori et al.~\cite{GNN_Gori} and Scarselli et al.~\cite{GNN_Scarselli} as a generalization
of recursive neural networks that can directly deal with a more general class of graphs~\cite{GAT}. Recently, the Graph Convolutional Networks (GCN) generalize the traditional convolution to the data of graph structures. Bruna et al.~\cite{GCN_Bruna} proposed a general graph convolution framework based on Graph Laplacian. And then, Defferrard et al.~\cite{GCN_Defferrard} proposed a Chebyshev polynomial-based method to optimize the processing of eigenvalue decomposition. Based on that, Kipf et al.~\cite{GCN_Kipf} simplified them by restricting the filters to operate in a 1-step neighborhood around each node. After that, GAT~\cite{GAT} was proposed to realize the adaptive allocation of neighbor weights based on the attention mechanisms~\cite{Attention}.

\section{Preliminaries}\label{sec3}
\subsection{Demand Prediction}
Following the definition in~\cite{AAAI2018} and~\cite{AAAI2017}, we denote $R=\{r_1,r_2,…,r_N\}$ as the set of all regions in which the number of orders needs to be predicted, where $N$ is the total number of regions. We divide an area of the city into rectangular regions according to their actual geographical (latitude and longitude) distribution, as the settings in~\cite{AAAI2018,Origin-Destination-Demand,AAAI2019}.

For temporal information, suppose each day can be segmented into $H$ time intervals and there are $D$ days in the dataset, we define $T=\{t_0,t_1,…,t_{H\times D-1}\}$ as the set of whole time intervals. Given the above definitions, we further formulate the following conceptions.

\textbf{Ride-hailing order:} A Ride-hailing order $o$ can be defined as $\langle o.r,o.t\rangle$ that contains the region $o.r$ where people call for the ride-hailing and the corresponding start time interval $o.t$.

\textbf{Ride-hailing demand:} The demand in one region $r$ and time interval $t$ is defined as the total ride-hailing orders during that time interval and location, which can be denoted as $x_r^t \in \mathbb{N}$. Therefore, we set $X^t \in \mathbb{N}^N$ as the demand of all regions in time interval $t$ with each element $x_r^t$ representing the demand of each region, and $N$ denotes the total number of regions.

\textbf{Demand Prediction:} The demand prediction problem aims to predict the data in the future one time step or several steps given the sequential data from the beginning to the current. We denote it as
\begin{equation}
    X^{t+1}= \mathcal{F}(\left\{X^{t-L+1},…,X^{t-1},X^{t}\right\})
\end{equation}\label{s}
where $L$ is the length of the input sequence.

We define the network of regions as a directed graph $G^t$ = ($V$,$E^t$,$A^t$) at time interval $t$, where $V$ is the set of nodes each of which denotes a single region, $E^t$ and $A^t$ represent the set of edges and the adjacency matrix of the graph $G^t$ respectively. As we mentioned above, the $E^t$ and $A^t$ are based on the real-time commuting relationships among regions at time interval $t$. Therefore, the demand prediction task in this work can be denoted as
\begin{equation}
    X^{t+1}= \mathcal{F}(\left\{(X^{t-L+1}, G^{t-L+1}),…,(X^{t-1}, G^{t-1}),(X^{t}, G^{t})\right\})
\end{equation}

\begin{figure}[t]
    \centering
    \includegraphics[width=.45\textwidth]{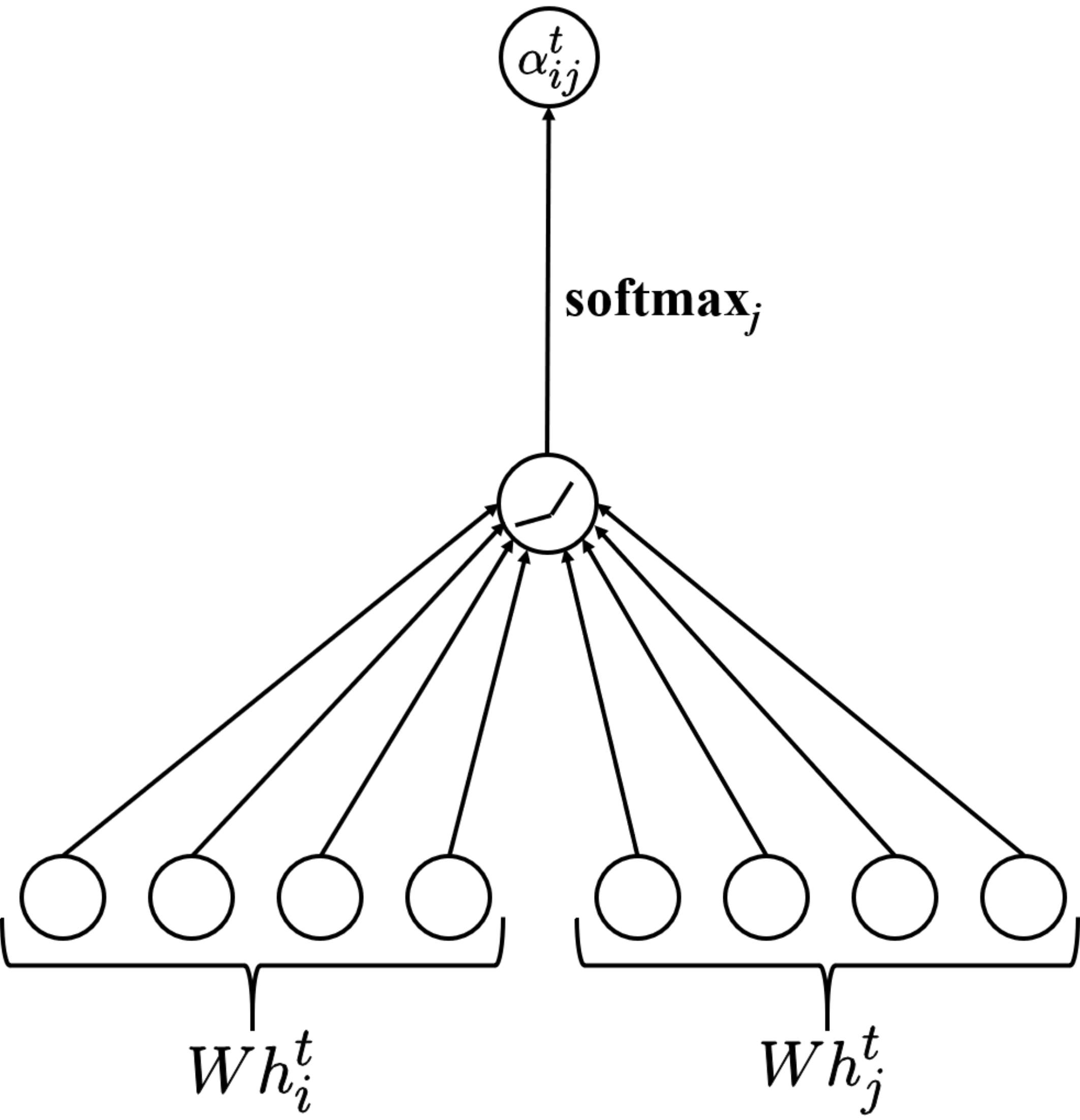}
    \caption{Graph attention layer~\cite{GAT}}
    \label{fig:attention_layer}
\end{figure}

\begin{figure*}[t]
    \centering
    \includegraphics[width=\textwidth]{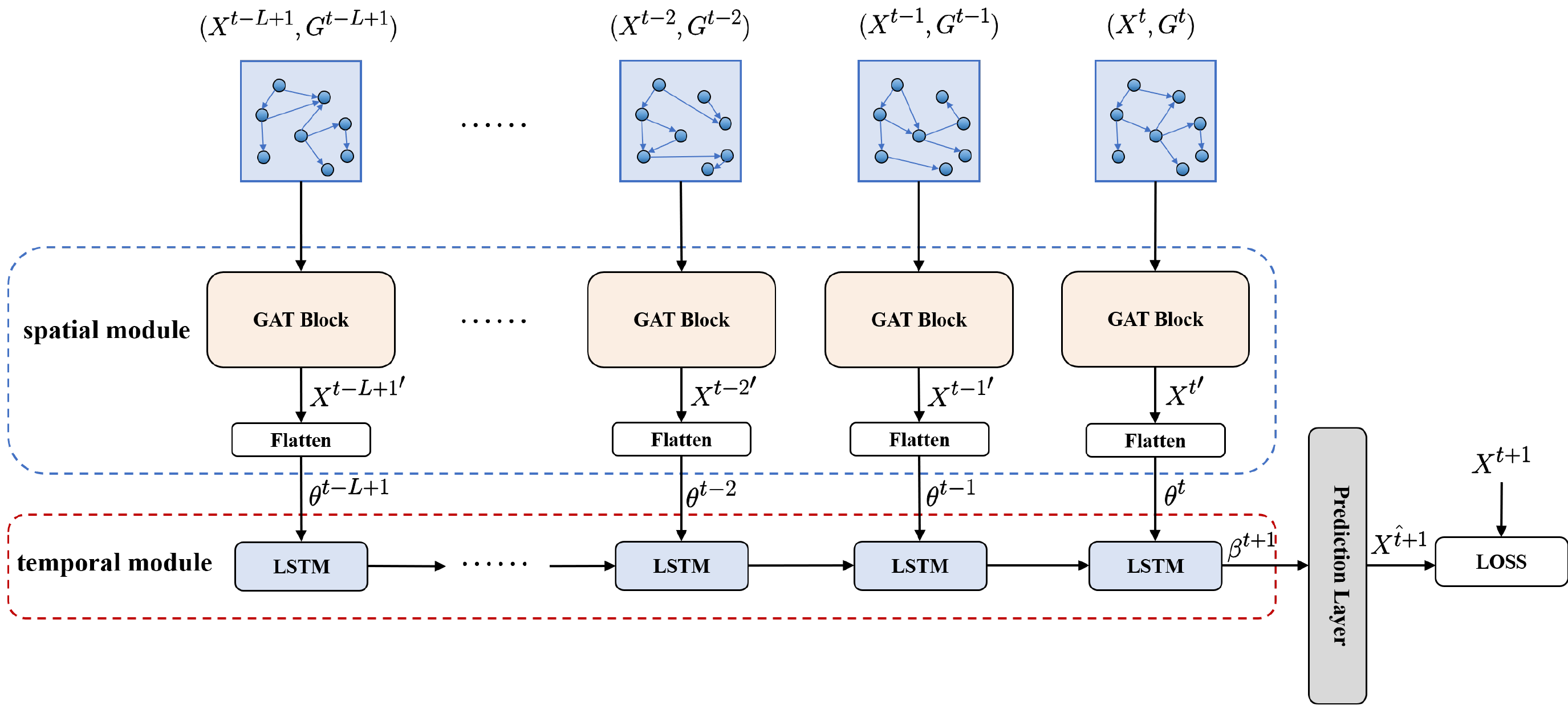}
    \caption{The Architecture of STDGAT. The spatial module uses a GAT Block to capture the spatial feature among regions. The GAT Block consists of graph attention layers. The temporal module applies an LSTM model to extract temporal information among the input sequence.}
    \label{fig:model}
\end{figure*}

\subsection{Graph Convolution Network}
Graph Convolution Network (GCN) is defined over a graph $G=(V,E,A)$ instead of applying convolution operation on regular grids, where $V$ and $E$ are the set of all nodes and the set of edges in the graph $G$, and $A\in \mathbb{R}^{|V|\times|V|}$ denotes the adjacency matrix of the graph $G$. GCN can overcome the shortage of traditional CNN that the lack of processing non-Euclidean structure data~\cite{AAAI2019}. Currently, one of the most popular GCN approaches to applying convolution operation over graph is introduced by Defferrard et al.~\cite{GCN_Defferrard} based on Chebyshev polynomials approximation, which is defined as:
\begin{equation}
        x' = \sigma\Bigg(\sum_{j=0}^{K-1}\alpha_{j}L^{j}x\Bigg)
\label{eqChebyshev}
\end{equation}
where $x$ is the input feature, $(\alpha_0, \alpha_1,...,\alpha_{K-1})$ is learnable coefficients, $L^j$ presents the $j$-th power graph Laplacian matrix, and $\sigma(\cdot)$ denotes the activation function.
In the equation~\ref{eqChebyshev}, the formulation of the graph Laplacian matrix can be presented as
\begin{equation}
        L = I - D^{-1/2}AD^{-1/2}
\end{equation}
where $I\in\mathbb{R}^{|V|\times|V|}$ is the identity matrix and $D$ denotes the degree matrix of the graph.

Based on the above GCN theory, Kipf et al.~\cite{GCN_Kipf} stacked multiple localized graph convolutional layers the first-order approximation of graph
Laplacian to define a layer-wise linear formulation~\cite{STGCN}. By this way, the formulation of the GCN layer can be denoted as:
\begin{equation}
        x' = \sigma(W\widetilde{D}^{-1/2}\widetilde{A}\widetilde{D}^{-1/2}x)
\end{equation}
where $W$ is the trainable parameters of the layer. $\widetilde{A}$ and $\widetilde{D}$ are the renormalized matrix that $\widetilde{A}=A+I$ and $\widetilde{D}_{ii}=\sum\nolimits_{j}\widetilde{A}_{ij}$. In this way, the GCN layer can reduce the learnable parameters which is highly efficient for large-scale graphs, since the order of the approximation is limited to one~\cite{STGCN}.

\subsection{Graph Attention Layer}
Graph Attention Network (GAT) is an advanced Graph Convolution Network (GCN) with attention mechanisms~\cite{Attention} that has been widely used in computer vision~\cite{Attention_using_CV1,Attention_using_CV2}, and natural language processing~\cite{Attention_using_NLP1,Attention_using_NLP2,Attention_using_NLP3}.

The graph attention layer is the base component of GAT, which is used to learn attention coefficients between node pairs and update the hidden feature of each node.

We first denote $v_i$ as the $i-$th node of the nodes set $V$, and then, due to the graph that inputted into the GAT layers is the time-specific graph structure as we described above, the connection relationships of the nodes are different at different time intervals. Based on this, we further denote $v_i^t$ as node $v_i$ that in time interval $t$. After that, We represent the feature of node $v_i^t$ in layer $l$ at time interval $t$ as $h_i^t\in \mathbb{R}^{d(l)}$, where $d(l)$ is the length of the feature of node $v_i^t$ in layer $l$. In the first graph attention layer, $h_i^t$ is the demand value of region $i$ at time interval $t$. As figure~\ref{fig:attention_layer} shows, the attention coefficient between node $v_i^t$ and its neighbor $v_j^t$ at time interval $t$ can be presented as
\begin{equation}
    e_{ij}^t = a(Wh_i^t,Wh_j^t)
\end{equation}
where $W\in \mathbb{R}^{d(l+1)\times d(l)}$ is the learnable parameters of layer $l$, and $a(\cdot)$ is the function that calculates the correlation between node $v_i^t$ and $v_j^t$. Note that, $v_j^t\in N_{v_i}^t$, where $ N_{v_i}^t$ denotes the set of the neighbor nodes of $v_i^t$. For $a(\cdot)$, we follow~\cite{GAT}, and use a trainale feedforward neural network, parametrized by a weight vector $\vec{a}\in\mathbb{R}^{2d(l+1)}$, which can denoted as
\begin{equation}
    e_{ij}^t = LeakyReLU(\vec{a}^T[Wh_i^t||Wh_j^t])
\end{equation}
where $\cdot^T$ and $||$ denote the transposition and concatenation operation respectively.

And then, we normalize the attention coefficient by a softmax function for easily comparable:
\begin{equation}
    \alpha_{ij}^t=softmax(e_{ij}^t)=\frac{exp(e_{ij}^t)}{\sum\nolimits_{v_k\in N_{v_i}^t}exp(e_{ik}^t)}
\end{equation}
After the computing of coefficients above, the new feature vector of $v_i^t$ can be computed according to the weighted summation of attention mechanism:
\begin{equation}
    {h_i^t}' = \sigma\Bigg(\sum_{v_j\in N_{v_i}^t}\alpha_{ij}^{t}Wh_j^t\Bigg)
\end{equation}
where $\sigma(\cdot)$ is the activation function.

Note that, the parameters of the graph attention layer $W$ and $\vec{a}$ are shared among the input sequence.

\section{Spatial-Temporal Dynamic Graph Attention Network}\label{sec4}
Figure~\ref{fig:model} shows the overall architecture of our model, which contains a spatial module, a temporal module and a prediction layer.

\subsection{Spatial Module}
\begin{figure}[t]
    \centering
    \includegraphics[width=.2\textwidth]{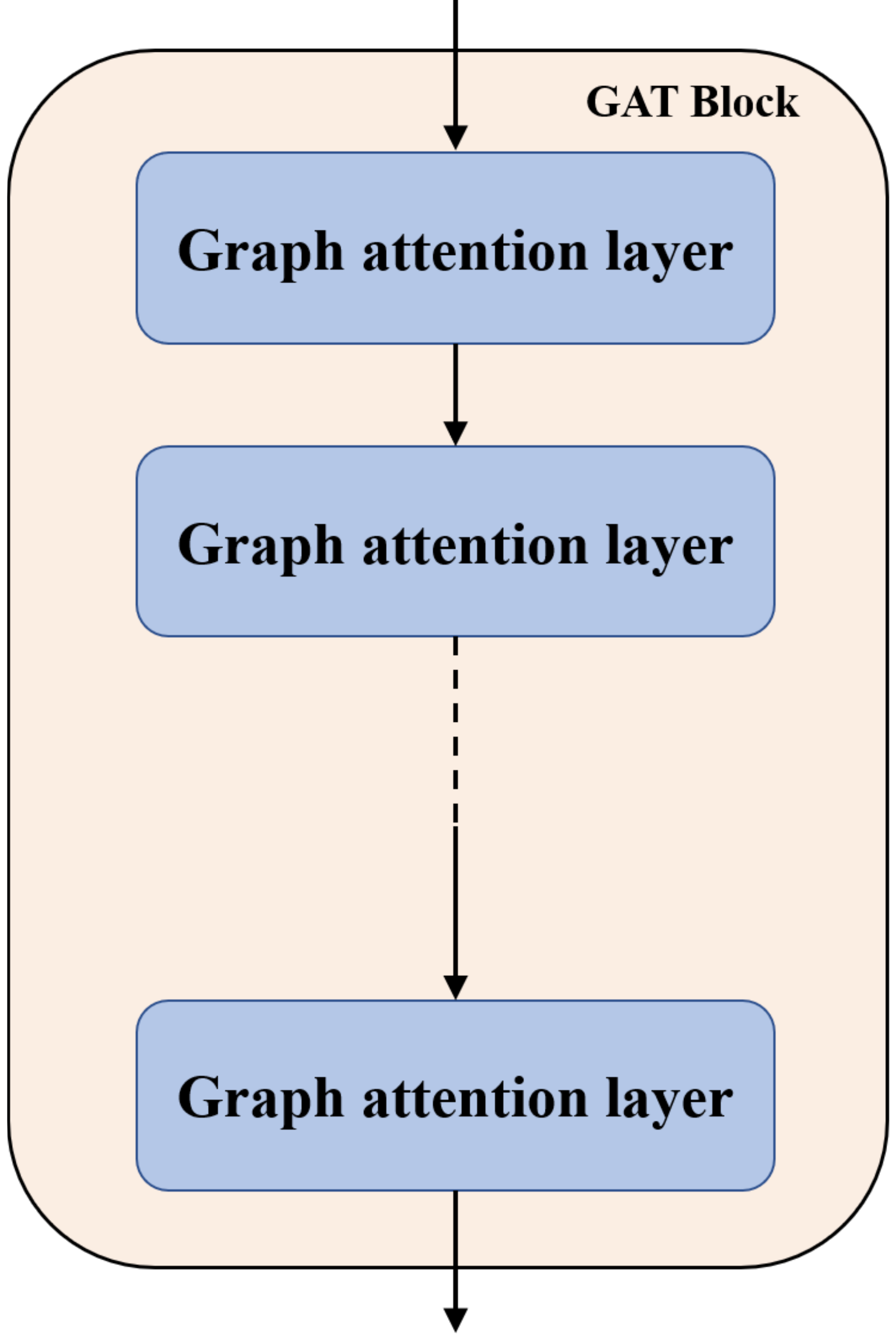}
    \caption{Internal structure of GAT Block}
    \label{fig:GAT_Block}
\end{figure}

The spatial module aims to extract the spatial correlations of the city at each time interval. The spatial module consisted of a GAT Block shared among the input sequence and a flatten layer. The internal structure of the GAT Block is shown in figure~\ref{fig:GAT_Block}. For each graph attention layer in GAT Block, we denote it as
\begin{equation}
    X_{l+1}^t = f_l(X_{l}^t)
\end{equation}
where $X_l^t\in\mathbb{R}^{N\times d(l)}$ is the input of the $l$-th layer of the GAT Block at time interval $t$, and $f_l$ denotes the graph attention operation of the $l$-th layer as we mentioned above. $N$ and $d(l)$ present the total number of regions and length of the feature of each node in layer $l$ respectively.

After all the graph attention layers, the output of the GAT Block can be presented as
\begin{equation}
    {X^t}' = \mathcal{F}(X^t)
\end{equation}
where ${X^t}'\in\mathbb{R}^{N\times d}$ is the spatial feature output from GAT Block at time interval $t$, and $\mathcal{F}(\cdot)$ denotes the overview operation of GAT Block, and $d$ is the length of the feature of each node after GAT Block.

After the graph attention operation, we apply a flatten layer to transform ${X_t}'$ that output from GAT Block to a feature vector $\theta^t\in\mathbb{R}^{Nd}$. The whole output $S_{t+1}\in\mathbb{R}^{L\times Nd}$ represents all the spatial features extracted from the input demand sequence through the spatial module, which can be denoted as
\begin{equation}
    S_{t+1} = [\theta^n|n=t,t-1,t-2,…,t-L+1]
\end{equation}

\subsection{Temporal Module}
Due to the demand sequence data is a kind of time series, we use a temporal module for modeling the temporal dependence of the demand sequence. In the sequence learning tasks, Recurrent Neural Networks (RNN) have achieved good results\cite{seq2seq}. The incorporation of Long Short-Term Memory (LSTM) overcomes the shortage of traditional recurrent networks that learning long-term dependencies is difficult~\cite{GradientFlowinRecurrentNets}. Moreover, some previous spatial-temporal prediction works~\cite{Origin-Destination-Demand,AAAI2018,AAAI2019} have proven the excellent performance of LSTM in processing such sequential data. Therefore, we use the LSTM network to model the temporal dependence of the demand sequential data in our temporal module.

Briefly speaking, LSTM introduces a memory cell $c_t$ to accumulate the previous sequence information. Specifically, at time $t$, given an input $x_t$, the LSTM uses an input gate $i_t$ and a forget gate $f_t$ to update its memory cell $c_t$, and uses an output gate $o_t$ to control the hidden state $h_t$. The formulation is defined as follows:
\begin{equation}
    \begin{split}
        &i_t=\sigma(W_{ii}x_t + b_{ii} + W_{hi}h_{t-1} + b_{hi})\\
        &f_t=\sigma(W_{if}x_t + b_{if} + W_{hf}h_{t-1} + b_{hf})\\
        &g_t=\tanh(W_{ig}x_t + b_{ig} + W_{hg}h_{t-1} + b_{hg})\\
        &o_t=\sigma(W_{io}x_t + b_{io} + W_{ho}h_{t-1} + b_{ho})\\
        &c_t=f_t\circ c_{t-1} + i_t\circ g_t\\
        &h_t=o_t\circ \tanh(c_t)
    \end{split}
\end{equation}
where $\circ$ denotes the Hadamard product, and $\sigma$ represents the sigmoid function. $W_{pq},b_{pq}(p\in(i,h),q\in(i,f,g,o))$ are the learnable parameters of the LSTM while $c_{t-1}$ and $h_{t-1}$ are the memory cell state and the hidden state at time $t-1$. Please refer to~\cite{LSTM,GradientFlowinRecurrentNets} for more details.

In our model, the LSTM net takes $S_{t+1}$ as input, which is the output of the spatial module. We use $\beta^{t+1}\in \mathbb{R}^k$ to represent the output of the LSTM net in our temporal module, where $k$ is the length of the output of the LSTM net.

\subsection{Prediction Layer}
After the above spatial and temporal module, we have captured the joint spatial-temporal feature of the input sequence. And then, we apply a prediction layer to map the extracted feature to a prediction demand vector. The prediction layer in our model is a learnable fully connected layer with $N$ neurons, where $N$ is the total number of regions. The formulation of the prediction layer can be expressed as follow:
\begin{equation}
    \hat{X^{t+1}} = f(W_{FC}\beta^{t+1}+b_{FC})
\end{equation}
where $\hat{X^{t+1}}\in \mathbb{R}^N$ present the demand prediction result, $\beta^{t+1}$ is the output from the temporal module, $W_{FC}$ and $b_{FC}$ are the weights and biases of the prediction layer respectively, and $f(\cdot)$ denotes the activation function.

\subsection{Implementation Details}
In the experiments, we set the length of the input sequence L to 5, and the total number of regions in our experiments is 121. In the spatial module, the number of graph attention layers in GAT Block is set to 3, each of which has 32 hidden units. In the temporal module, the LSTM net has 1 hidden layer with 512 neurons. All the activation functions used in graph attention layers are $LeakyReLU$, and that used in the prediction layer is $ReLU$. We optimize our proposed model via Adam~\cite{Adam} optimization by minimizing the Mean Squared Error (MSE) loss between the prediction results and the ground truths. In the experiments we conducted, the learning rate and the weight decay are set to $10^{-3}$ and $5e-5$ respectively. For the training data, 80\% of it is selected for training and the remaining 20\% is chosen as the validation set. We implement our network with Pytorch~\cite{pytorch} and train it for 200 epochs on 2 NVIDIA 1080Ti GPUs.

\section{Experiments}\label{sec5}
\subsection{Dataset}
In this paper, we use the ride-hailing dataset from Didi Chuxing, which is one of the largest online ride-hailing companies in China. The dataset contains ride-hailing orders from May 1st to Oct. 31st in the city of Haikou. There are 121 regions in the dataset, and the size of each is about $1km \times 1km$ according to industrial practice~\cite{AAAI2019}. The total number of orders and time intervals in the dataset are 12,185,427 and 4,416 respectively.

In our experiments, the data from 05/01/2017 to 09/30/2017 is used for training, and the remaining data (from 10/01/2017 to 10/31/2017) is for testing. For the training data, we select 80\% of it for training, and the remaining 20\% is chosen as the validation set. 

\begin{table}
    \centering
    \caption{The description of the dataset}
    \begin{tabular}{c|c}
         \hline
         Dataset & Didi Chuxing\\
        \hline
         Location & Haikou\\
         Time span & 5/1/2017 – 10/31/2017\\
         Time interval & 1 hour\\
         Available time interval & 4,416\\
         Total number of regions & 121\\
         Grid size & $1km \times 1km$\\
        \hline 
    \end{tabular}
    \label{tab:dataset}
\end{table}

\subsection{Loss Function}
In this section, we describe the loss function we used in the training step of our experiments. 

We use Mean Square Error (MSE) Loss as the loss function, and train our model by minimizing it, which is defined as:
\begin{equation}
    \mathcal{L}(\Theta)=\frac{1}{z}\sum_{i}(y_i-\hat{y_i})^2
\end{equation}
where $\Theta$ are all the learnable parameters in our proposed model, $\hat{y_i}$ and ${y_i}$ denote the predicted value and ground truth respectively, and ${z}$ is the number of all samples. 

\subsection{Evaluation Metric}
In our experiments, we adopt the Rooted Mean Square Error (RMSE), Mean Average Percentage Error (MAPE) and Mean Absolute Error (MAE) as the metrics to evaluate the performance of all methods, which are defined as follows:

\begin{equation}
    RMSE=\sqrt{\frac{1}{z}\sum_{i=1}^{z}(y_i-\hat{y_i})^2}
\end{equation}

\begin{equation}
    MAPE=\frac{1}{z}\sum_{i=1}^{z}\frac{|y_i-\hat{y_i}|}{y_i}
\end{equation}

\begin{equation}
    MAE=\frac{1}{z}\sum_{i=1}^{z}|y_i-\hat{y_i}|
\end{equation}
where $\hat{y_i}$ and ${y_i}$ denote the predicted value and ground truth respectively, and ${z}$ is the number of all samples. 

\subsection{Baselines}
We compared our proposed model with the following methods. We tune the parameters of all methods and report their best performance.
\begin{itemize}
    \item \textbf{Historical average (HA)}: Historical average predicts the future demand by averaging the historical demands at the location given in the same relative time interval.
    \item \textbf{Autoregressive integrated moving average (ARIMA)}: Auto-Regression Integrated Moving Average (ARIMA) is a well-known model used for time series prediction.
    \item \textbf{Lasso regression (Lasso)}: Lasso regression is a linear regression method with $L_1$ regularization.
    \item \textbf{Ridge regression (Ridge)}: Ridge regression is a linear regression method with $L_2$ regularization.
    \item \textbf{XGBoost}~\cite{XGBoost}: XGBoost is a powerful boosting trees based method that is widely used in many data mining tasks.
    \item \textbf{Multiple layer perception (MLP)}: MLP is a neural network consisted of four hidden fully connected layers with 128, 128, 64, 64 neurons respectively.
    \item \textbf{DMVST-Net}~\cite{AAAI2018}: DMVST-Net is a deep learning model based on CNN and LSTM for taxi demand prediction. It also contains a graph embedding module to capture similar demand patterns among regions.
    \item \textbf{STGCN}~\cite{STGCN}: STGCN is a GCN-based model for traffic forecasting, which uses GCN to extract spatial correlations and 1-d convolutional kernel to model temporal dependencies.
    \item \textbf{ST-MGCN}~\cite{AAAI2019}: ST-MGCN is a state-of-the-art deep learning method for ride-hailing demand prediction. The ST-MGCN uses neighborhood, functional similarity, and transportation connectivity to construct multi-graphs to extract spatial features among regions. And the Contextual Gated Recurrent Neural Network (CGRNN) is proposed to capture temporal dependencies.  
\end{itemize}

\begin{figure*}[t]
    \centering
    \includegraphics[width=.85\textwidth]{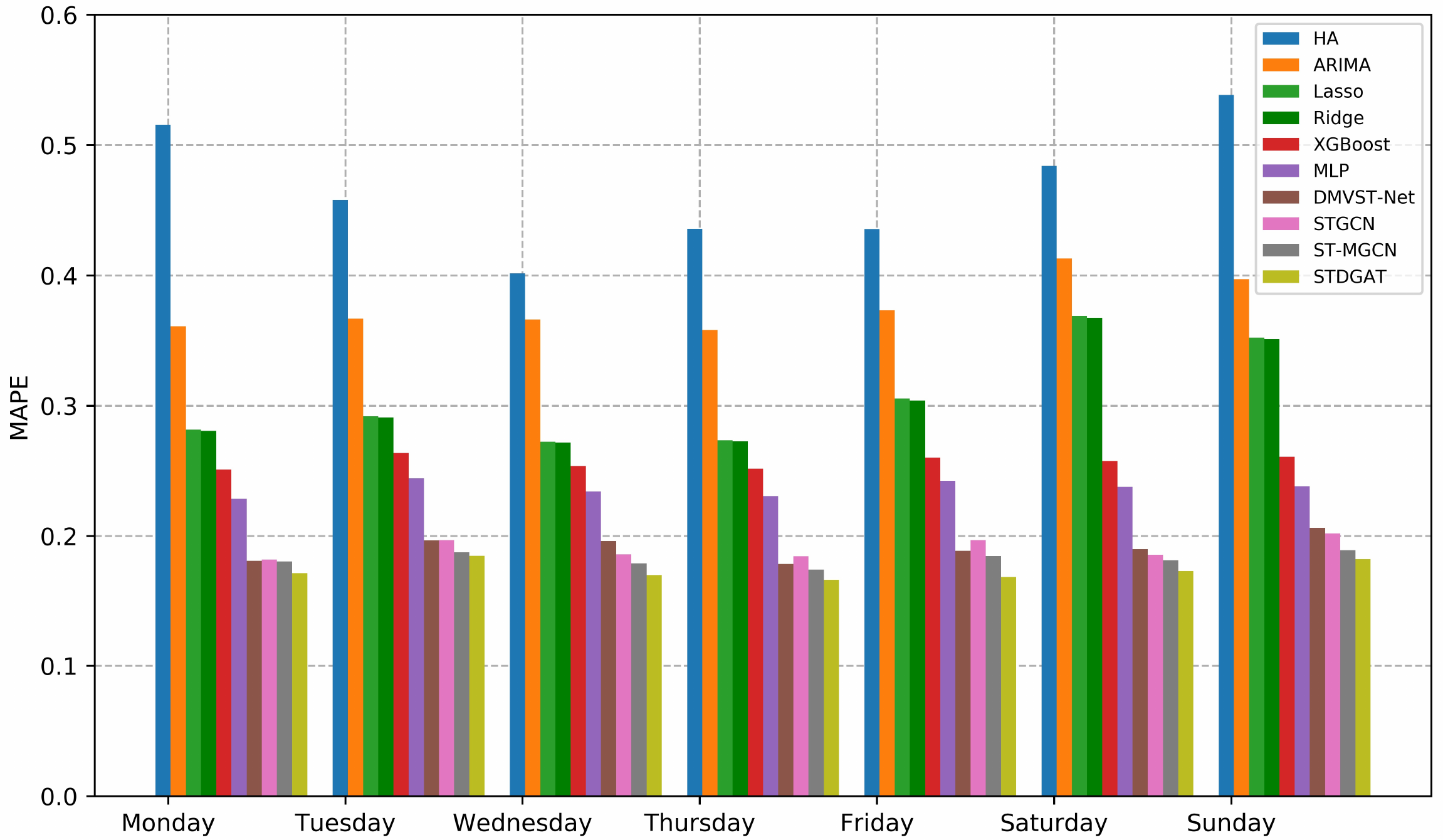}
    \caption{Performance on Different Days}
    \label{fig:different_days}
\end{figure*}

\subsection{Comparison with Baselines}
Table~\ref{tab:baselines} shows the performance of our proposed model and compared baselines. Our proposed STDGAT achieves the lowest RMSE (7.8811), lowest MAPE (0.1744) and MAE (4.0881) among all methods. More specifically, we can see that the HA and ARIMA perform poorly, which have MAPE of 0.4710 and 0.3769 respectively, as they only rely on historical values for prediction without the extraction of other related features. The linear regression methods, i.e. Lasso and Ridge, perform better than the above two methods due to the consideration of more context relationships among sequence. However, the linear regression methods do not extract more spatial or temporal features for prediction. The XGBoost and MLP further extract more hidden information from the sequence, which improves their performance significantly. However, all the above methods lacks modeling for spatial and temporal dependencies

The deep learning methods DMVST-Net, STGCN, and ST-MGCN further consider the spatial and temporal dependencies. Compared with these three methods, our proposed STDGAT achieves better performance by considering the adaptive pair-wise correlations for different neighbor regions and the time-specific commuting-based dynamic spatial modeling. Specifically, DMVST-Net just embeds a static graph into a vector when extracting global spatial features, which has a limited effect on spatial modeling. The STGCN and ST-MGCN model the non-Euclidean correlations among regions, which has been proven to be reasonable and necessary. However, due to applying traditional GCN, they only allocate the same pair-wise correlations for different neighbor regions, which fails to assign different importance to different neighbor regions. Moreover, the time-specific spatial correlation is being ignored in these methods. Our STDGAT further addresses these two issues by proposing the GAT-based spatial modeling and the time-specific commuting-based dynamic graph structure that further achieves better performance.

\begin{table}
    \centering
    \caption{Comparison with Different Baselines.}
    \begin{tabular}{p{50pt}<{\centering}|p{48pt}<{\centering}|p{48pt}<{\centering}|p{48pt}<{\centering}}
    \toprule
    Method & RMSE & MAPE & MAE\\
    \hline
    \hline
    HA & 24.1629 & 0.4710 & 13.5835\\
    ARIMA & 22.5440 & 0.3769 & 12.4762\\
    Lasso & 19.5851 & 0.3080 & 11.3082\\
    Ridge & 19.5968 & 0.3069 & 11.3178\\
    XGBoost & 11.3143 & 0.2572 & 6.5740\\
    MLP & 10.4753 & 0.2367 & 5.6806\\
    \hline
    DMVST-Net & 8.8695 & 0.1914 & 4.5365\\
    STGCN & 8.6153 & 0.1909 & 4.5579\\
    ST-MGCN & 8.5190 & 0.1827 & 4.4022\\
    \hline
    \hline
    STDGAT & \textbf{7.8811} & \textbf{0.1744} & \textbf{4.0881}\\
    \bottomrule
    \end{tabular}
    \label{tab:baselines}
\end{table}

\subsection{Performance on Different Days}
In this section we compare the performance of different methods on different days of the week. As figure~\ref{fig:different_days} shows, our proposed STDGAT consistently outperforms other compared methods in all seven days, which illustrates the robust of our proposed STDGAT. Furthermore, we also evaluate the performance of all the methods on weekdays and weekends respectively. The results have been shown in table~\ref{tab:weekdays_weekends}, and from which we can see that our proposed STDGAT still achieves the best performance on both weekdays and weekends. 

Besides, we can see that the prediction results of all the methods on weekends are worth than that on weekdays. The reason for this phenomenon has been found by Yao et al.~\cite{AAAI2018} that the demand patterns of taxis on weekends are less regular than that on weekdays. And from these results, we can conclude that the regular demand patterns are easier for models to learn.

\begin{table}
    \centering
    \caption{The MAPE of Different Methods on Weekdays and Weekends.}
    \begin{tabular}{p{50pt}<{\centering}|p{50pt}<{\centering}|p{50pt}<{\centering}}
    \toprule
    Method & Weekdays & Weekends\\
    \hline
    \hline
    HA & 0.4527 & 0.5132\\
    ARIMA & 0.3650 & 0.4044\\
    Lasso & 0.2854 & 0.3600\\
    Ridge & 0.2844 & 0.3586\\
    XGBoost & 0.2563 & 0.2593\\
    MLP & 0.2362 & 0.2381\\
    DMVST-Net & 0.1882 & 0.1987\\
    STGCN & 0.1894 & 0.1944\\
    ST-MGCN & 0.1815 & 0.1855\\
    STDGAT & \textbf{0.1728} & \textbf{0.1780}\\
    \bottomrule
    \end{tabular}
    \label{tab:weekdays_weekends}
\end{table}

\subsection{Comparison with Variants of Our Model}
Our model consists of three components (spatial module, temporal module, and prediction layer). To explore the influence of different modules, we divide them and implement the following networks:
\begin{itemize}
    \item \textbf{spatial module + prediction layer}: This network is the combination of the spatial module and the prediction layer of our complete model. In this model, only spatial features are extracted for prediction.
    \item \textbf{temporal module + prediction layer}: In this network, we use the temporal module of our proposed model to capture the temporal dependencies among the input sequence, and the prediction layer is used to output the final result.
    \item \textbf{STDGAT}: Our proposed model, which captures joint spatial-temporal information to predict the final results.
\end{itemize}

\begin{table}
    \footnotesize
    \centering
    \caption{Comparison with Different Modules.}
    \begin{tabular}{p{120pt}<{\centering}|p{25pt}<{\centering}|p{25pt}<{\centering}|p{25pt}<{\centering}}
    \toprule
    Method & RMSE & MAPE & MAE\\
    \hline
    \hline
    spatial module + prediction layer & 9.6273 & 0.2143 & 5.2131\\
    temporal module + prediction layer & 9.3151 & 0.1946& 4.5964\\
    STDGAT & \textbf{7.8811} & \textbf{0.1744} & \textbf{4.0881}\\
    \bottomrule
    \end{tabular}
    \label{tab:different_modules}
\end{table}

\begin{figure*}[t]
\centering
\subfloat[]{
\centering
\includegraphics[width=0.45\textwidth]{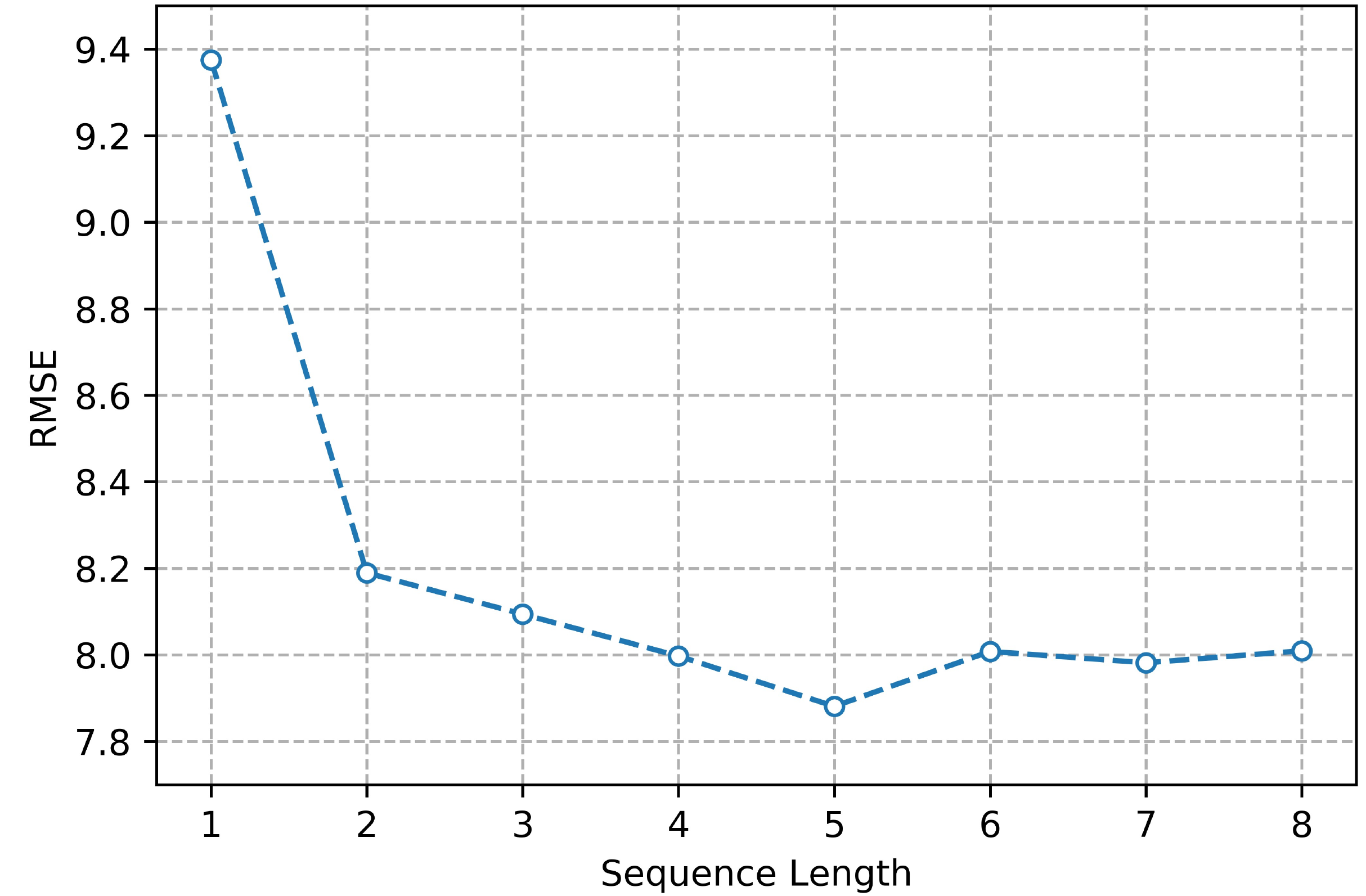}
\label{fig:Num_sequence_length}
}
\subfloat[]{
\centering
\includegraphics[width=0.45\textwidth]{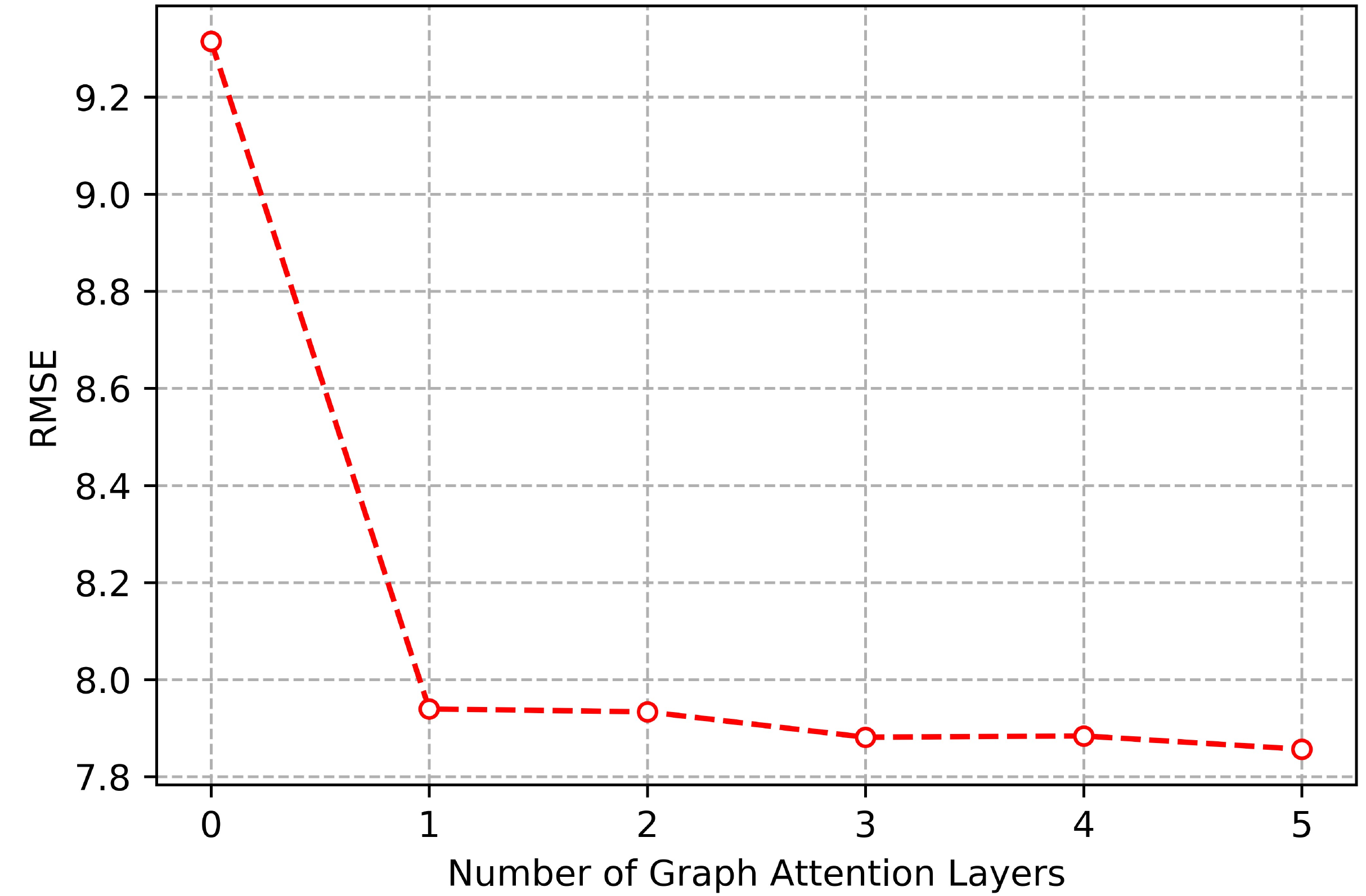}
\label{fig:Num_attention_layers}
}
\centering
\caption{(a) RMSE with respect to the length of the input sequence. (b) RMSE with respect to the number of graph attention layers.}
\label{fig:seqAndatt}
\end{figure*}

Table~\ref{tab:different_modules} shows the testing results of different modules. We can see that the RMSE, MAPE, and MAE of the spatial module + prediction layer are 9.6273, 0.2143, and 5.2131 respectively, and that of the temporal module + prediction layer are 9.3151 (RMSE), 0.1946 (MAPE), and 4.5964 (MAE). The performance of our complete model is better than separate module, which means modeling joint spatial-temporal dependencies can perform better than a separate one.

The above experiments show that our proposed model achieves a good result in the demand prediction task. However, we also need to prove the rationality and the performance of our proposed time-specific commuting-based graph attention mode. To evaluate our view, we construct the following variant of our proposed model:
\begin{itemize}
    \item \textbf{STDGAT-fixed}: This network is a variant of our proposed model, which uses a fixed graph structure in different time intervals based on the geographical neighborhood.
\end{itemize}

\begin{table}
    \centering
    \caption{Comparison with Variants of Our Model.}
    \begin{tabular}{p{80pt}<{\centering}|p{30pt}<{\centering}|p{30pt}<{\centering}|p{30pt}<{\centering}}
    \toprule
    Method & RMSE & MAPE & MAE\\
    \hline
    \hline
    STDGAT-fixed & 8.2334 &  0.1788 & 4.1917\\
    STDGAT & \textbf{7.8811} & \textbf{0.1744} & \textbf{4.0881}\\
    \bottomrule
    \end{tabular}
    \label{tab:variants}
\end{table}

As shown in table~\ref{tab:variants}, our STDGAT outperforms the STDGAT-fixed, that means, our proposed time-specific commuting-based graph attention mode has a better performance than traditional geographical neighborhood-based graph representation learning on the ride-hailing demand prediction task.

\subsection{Influence of Sequence Length and Number of GAT Layers}
In this section, we explore the influence of the length of the input sequence and the influence of the number of graph attention layers.

Figure~\ref{fig:Num_sequence_length} shows the prediction results of different input sequence length. We can see that our method achieves the best performance when sequence length is set to 5. The prediction error decreases with the increasing sequence length from 1 to 4, which means temporal dependency plays an important role in the task. However, as the length of the sequence increases to more than 5 hours, the performance of our model slightly degrades and it has a fluctuation. One potential reason is that with the length of the input sequence growing, many more parameters need to be learned, which makes the training harder.

In figure~\ref{fig:Num_attention_layers}, we show the performance of our model with respect to the number of graph attention layers. We can see that the prediction error decreases as the number of graph attention layers growing from 0 to 5. That means, with the number of graph attention layers rising, the performance of our model becomes better. This due to the fact that the original features are aggregated with their neighbors as layers deepen, which makes deeper layers have larger receptive fields. As we know, larger receptive fields can capture more spatial correlations. Therefore, the model can learn more spatial information as layers deepen to improve its performance.

\section{Conclusion and Future Work}\label{sec5}
In this paper, we propose a novel deep learning-based method for ride-hailing demand prediction. We apply the Graph Attention Network (GAT) to extract the non-Euclidean correlations among regions, which achieves different pair-wise correlations by adaptive importance allocation for different neighbor regions. Furthermore, we propose a time-specific commuting-based graph attention mode that allows the model to capture the dynamic spatial relationships throughout the timeline. We evaluate our model on a large-scale ride-hailing dataset from Didi Chuxing, and the experimental results show that our STDGAT significantly outperforms the state-of-the-art baselines. In the future, we will evaluate our proposed method on other spatial-temporal prediction tasks. And we will consider some other features to further improve the performance of our model, such as meteorology data, holiday data. 

\section{Acknowledgments}
This work was supported in part by the National Key Research and Development Project under grant 2019YFB1706101, in part by the Science-Technology Foundation of Chongqing, China under grant cstc2019jscx-mbdx0083. Data source: Didi Chuxing GAIA Initiative.

\bibliography{paper}

\end{document}